\documentclass{llncs} 
\usepackage{graphicx}
\begin{document}

\title{Liquid State Genetic Programming}

\author{Mihai Oltean}
\institute{Department of Computer Science\\
Faculty of Mathematics and Computer Science\\
Babe\c s-Bolyai University, Kog\u alniceanu 1\\
Cluj-Napoca, 3400, Romania.
\\
\email{mihai.oltean@gmail.com}\\
}

\maketitle

\begin{abstract}

A new Genetic Programming variant called Liquid State Genetic Programming (LSGP) is proposed in this paper. LSGP is a hybrid method combining a dynamic memory for storing the inputs (the liquid) and a Genetic Programming technique used for the problem solving part. Several numerical experiments with LSGP are performed by using several benchmarking problems. Numerical experiments show that LSGP performs similarly and sometimes even better than standard Genetic Programming for the considered test problems.

\end{abstract}

\section{Introduction}

Liquid State Machine (LSM) is a technique recently described in the literature \cite{mass1}, \cite{mass2}. It
provides a theoretical framework for a neural network that can process signals presented
temporally. The system is comprised of two subcomponents, the \textit{liquid} and the
\textit{readout}. The former acts as a decaying memory, while the latter acts as the main
pattern recognition unit.

Liquid State Genetic Programming (LSGP), introduced in this paper, is similar to both LSM and Genetic Programming (GP) \cite{koza1} as it uses a dynamic memory (the liquid) and a GP algorithm which is the actual problem solver. The liquid is simulated by using some operations performed on the inputs. The purpose of the liquid is to transform the inputs into a form which can be more easily processed by the problem solver (GP). The liquid acts as some kind of preprocessor which combines the inputs using the standard functions available for the internal nodes of GP trees.

We have applied the LSGP on several test problems. Due to the space limitation we will present the results only for one difficult problem: even-parity. We choose to apply the proposed LSGP technique to the even-parity problems because according to Koza \cite{koza1} these problems appear to be the most difficult Boolean functions to be detected via a blind random search. 

Evolutionary techniques have been extensively used for evolving digital circuits \cite{koza1}, \cite{miller1}, due to their practical importance. The case of even-parity circuits was deeply analyzed \cite{koza1}, \cite{miller1} due to their simple representation. Standard GP was able to solve up to even-5 parity \cite{koza1}. Using the proposed LSGP we are able to easily solve up to even-8 parity problem. 

Numerical experiments, performed in this paper, show that LSGP performs similarly and sometimes even better than standard GP for the considered test problems. 

The paper is organized as follows: Liquid State Machines are briefly described in Sect. \ref{liquid_state_machines}. In Sect. \ref{lsgp} the proposed Liquid State Genetic Programming technique is described. The results of the numerical experiments for are presented in Sect. \ref{results}. Conclusions and further work directions are given in Sect. \ref{conclusions}.

\section{Liquid State Machines - a Brief Introduction}
\label{liquid_state_machines}

Liquid computing is a technique recently described in the literature \cite{mass1}, \cite{mass2}. It
provides a theoretical framework for an Artificial Neural Network (ANN) that can process signals presented
temporally. The system is comprised of two subcomponents, the \textit{liquid} and the
\textit{readout}. The former acts as a decaying memory, while the latter acts as the main
pattern recognition unit. To understand the system, a simple analogy
is used. If a small rock thrown in a cup (pool) of water it will generate ripples that are mathematically related both to characteristics of the rock,
as well as characteristics of the pool at the moment that the rock was thrown in. A camera
takes still images of the water's ripple patterns. A computer then analyzes these still
pictures. The result is that the computer should know something about the rock that was
thrown in. For example, it should know about how long ago the rock was thrown. The rock
represents a single bit from an input stream, or an action potential. The water is the liquid
memory. The computer functions as the readout.

Translated into ANN's language the idea behind Liquid State Machines has been implemented \cite{mass1}, \cite{mass2} as follows: Two ANNs are used: one of them plays the role of the liquid and the other is the actual solver. The inputs of the first network are the problem inputs and this network will reshape (modify) the inputs in order to be more easily handled by the second network. This second network, which is the actual problem solver, takes the inputs from some of the nodes (randomly chosen) of the first network. In this way, it is hopped that the structure of the second network (the actual problem solver) is simpler than the case when a single network is used for solving the problem.

\section{Liquid State Genetic Programming}
\label{lsgp}

In this section the proposed LSGP technique is described. LSGP is a hybrid method combining a technique for manipulating the liquid and a GP technique for the individuals.

\begin{figure*}[htbp]
\centerline{\includegraphics[width=4in,height=4.48in]{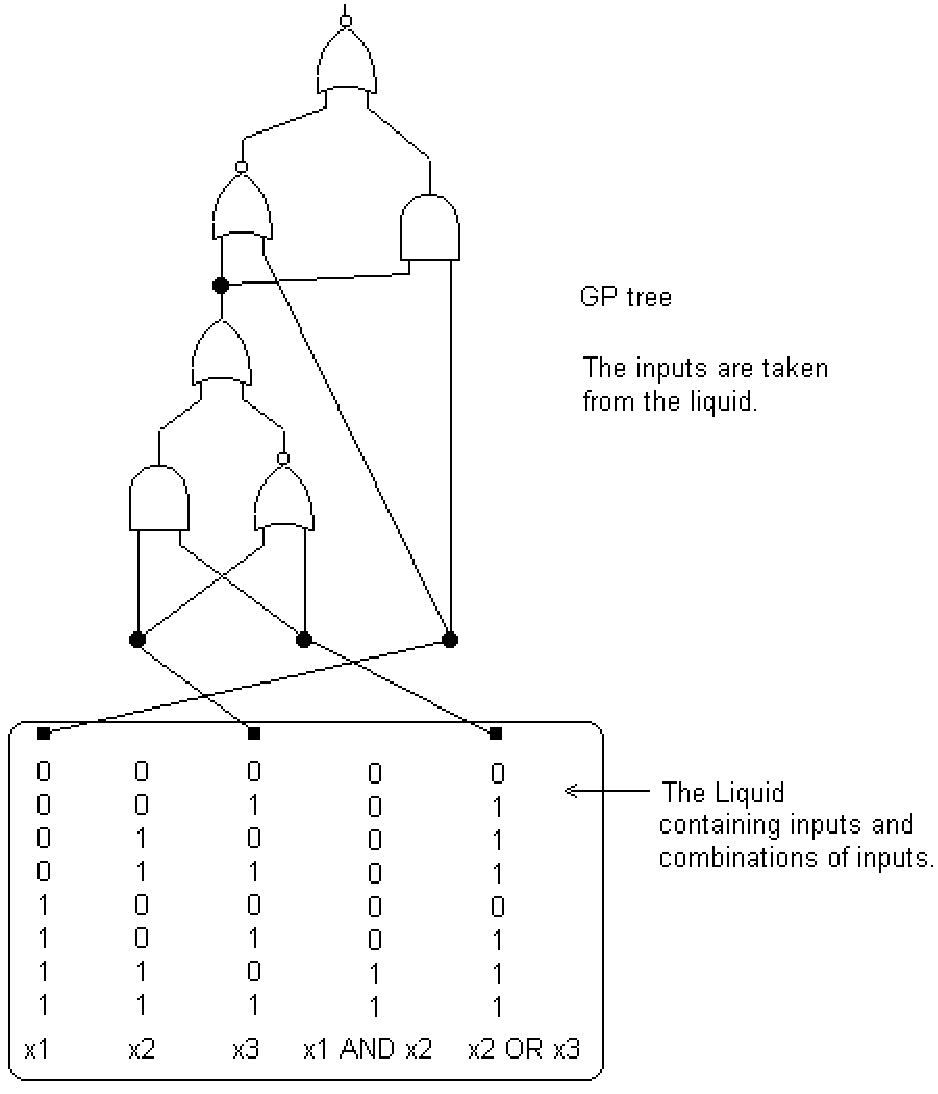}}
\caption{A \textit{liquid} and a GP program for the even-3-parity problem. The \textit{liquid} contains 5 items. Three of them are the standard inputs for the even-3-parity problem and 2 of them are combinations of the standard inputs. The inputs for the GP program are taken from the \textit{liquid}. The gates used by the GP program are the standard one: AND, OR, NAND, NOR}
\label{fig_gp_liquid}
\end{figure*}

For a better understanding we will start with a short example on how a LSGP individual looks like for the even-parity problem. The example is depicted in Fig. \ref{fig_gp_liquid}.

The liquid can be viewed as a set (a pool) of items (or individuals) which are subject to some strict rules which will be deeply explained in the next sections. The liquid is simulated by using some operations performed on the inputs. The purpose of the \textit{liquid} is to transform the inputs into a form which can be more easily processed by the problem solver (GP). The liquid acts as some kind of preprocessor which combines the inputs using the standard functions available for the internal nodes of GP trees.

The liquid and its accompanying rules can also be viewed as a simple GP algorithm that manipulates only the output of the tree rather than the entire tree. The state of the liquid will also evolve during the search process.

The GP algorithm is a standard one \cite{koza1} and due to the space limitations will not be detailed in this paper.

\subsection{Prerequisite}

The quality of a GP individual is usually computed using a set of fitness cases \cite{koza1}. For instance, the aim of symbolic regression is to find a mathematical expression that satisfies a set of $m$ fitness cases. 

We consider a problem with $n$ inputs: $x_{1}$, $x_{2}$, \ldots $x_{n}$ and one output $f$. The inputs are also called terminals \cite{koza1}. The function symbols that we use for constructing a mathematical expression are $F=\{+,-,*,/, sin\}$.

Each fitness case is given as an array of ($n+1$) real values:\\

\[
v_1^k ,v_2^k ,v_3^k ,...,v_n^k ,f_k 
\]

\noindent
where $v_j^k $ is the value of the $j^{th}$ attribute (which is $x_{j})$ in 
the $k^{th}$ fitness case and $f_{k}$ is the output for the $k^{th}$ fitness 
case.

Usually more fitness cases are given (denoted by $m$) and the task is to find the expression that best satisfies all these fitness cases. This is usually done by minimizing the quantity:

\[
Q=\sum\limits_{k = 1}^m {\left| {f_k - o_k } \right|} ,
\]

\noindent
where $f_{k}$ is the target value for the $k^{th}$ fitness case and $o_{k}$ is 
the actual (obtained) value for the $k^{th}$ fitness case.

\subsection{Representation of Liquid's Items}

Each individual (or item) in the liquid represents a mathematical expression obtained so far, but 
this individual does not explicitly store this expression. Each individual in the liquid stores only the obtained value, so far, for each fitness case. Thus an individual in the liquid is an array of values:\\

\[
(o_{1}, o_{2}, o_{3}, \ldots , o_{m})^{T},
\]

\noindent
where $o_{k}$ is the current value for the $k^{th}$ fitness case and $()^T$ is the notation for the transposed array. Each 
position in this array (a value $o_{k})$ is a gene. As we said it before 
behind these values is a mathematical expression whose evaluation has 
generated these values. However, we do not store this expression. We store 
only the values $o_{k}$.\\

\subsection{Initial Liquid}

The initial liquid contains individuals (items) whose values have been generated 
by simple expressions (made up by a single terminal). For instance, if an 
individual in the initial liquid represents the expression:\\

\[
E=x_{1},
\]

\noindent
then the corresponding individual in the liquid is represented as:\\

\[
C = (v_1^1 ,v_1^2 ,v_1^3 ,...,v_1^m)^{T}
\]

\noindent
where $v_j^k $ has been previously explained.\\

\begin{example}

For the particular case of the even-3-parity problem we have 3 inputs $x_1, x_2, x_3$ (see Fig. \ref{fig_gp_liquid}) and $2^3 = 8$ fitness cases which are listed in Table \ref{even3parity}:

\begin{table*}[htbp]
\caption{The truth table for the even-3-parity problem}
\label{even3parity}
\begin{center}
\begin{tabular}
{p{20pt}p{20pt}p{20pt}p{50pt}}
\hline
\textbf{$x_1$} & 
\textbf{$x_2$} & 
\textbf{$x_3$} & 
\textbf{Output} \\
\hline
0& 
0&
0& 
1\\
0& 
0&
1&
0 \\
0& 
1&
0&
0 \\
0& 
1&
1&
1 \\
1& 
0&
0&
0 \\
1& 
0&
1&
1 \\
1& 
1&
0&
1 \\
1& 
1&
1&
0 \\

\hline
\end{tabular}
\end{center}
\end{table*}

Each item in the liquid is an array with 8 values (one for each fitness case). There are only 3 possible items for initial liquid: $(00001111)^T$, $(00110011)^T$ and $(01010101)^T$ corresponding to the values for variables $x_1$, $x_2$ and $x_3$. Other items can appear in the liquid later as effect of the specific genetic operators (see Sect. \ref{liquidoperators}). Of course, multiple copies of the same item are allowed in a liquid at any moment of time.

\end{example}

It is desirable, but not necessary to have each variable represented at least once in the initial liquid. This means that the number of items in the liquid should be larger than the number of inputs of the problem being solved. However, an input can be inserted later as effect of the insertion operator (see Sect. \ref{insertion}).

\subsection{Operators Utilized for Modifying the Liquid}
\label{liquidoperators}

In this section the operators used for the Liquid part of the LSGP are 
described. Two operators are used: combination and insertion. These operators are specially designed for the liquid part of the proposed LSGP technique.

\subsubsection{Recombination}
\label{recombination}

The recombination operator is the only variation operator that creates new items in the liquid. For recombination several items (the parents) and a function symbol are selected. The offspring is obtained by applying the selected operator for each of the symbols of the parents.

The number of parents selected for combination depends on the number of 
arguments required by the selected function symbol. Two parents have to be selected for combination if the function symbol is a binary operator. A single parent needs to be selected if the function symbol is a unary operator.\\

\begin{example}

Let us suppose that the operator AND is selected. In this case two parents (items in the liquid): \\

\[
C_{1} = (p_{1}, p_{2}, \ldots , p_{m})^{T} and
\]

\[
C_{2} = (q_{1}, q_{2}, \ldots , q_{m})^{T}
\]

\noindent
are selected and the offspring $O$ is obtained as follows:

\[
O = (p_{1} AND q_{1}, p_{2} AND q_{2},\ldots , p_{m} AND q_{m})^{T}.
\]

\end{example}

\begin{example}

Let us suppose that the operator NOT is selected. In this case one parent (item in the liquid):

\[
C_{1} = (p_{1}, p_{2}, \ldots , p_{m})^{T} 
\]

\noindent
is selected and the offspring $O$ is obtained as follows:

\[
O = (NOT(p_{1}), NOT(p_{2}),\ldots , NOT(p_{m}))^{T}.
\]

\end{example}

\begin{remark}
The operators used for combining genes of the items in the liquid must be restricted to those used by the main GP algorithm. For instance, if the function set allowed in the main GP program is $F$ = \{AND, OR\}, then for the recombination part of the liquid we can use only these 2 operators. We cannot use other functions such as NOT, XOR etc.
\end{remark}

\subsubsection{Insertion}\label{insertion}

This operator inserts a simple expression (made up of a single terminal) in the liquid. This operator is useful when the liquid contains items representing very complex expressions that cannot improve the search. By inserting simple expressions we give a chance to the evolutionary process to choose another direction for evolution.

\subsection{The LSGP Algorithm}

Due to the special representation and due to the special operators, LSGP uses a special generational algorithm which is given below.

The LSGP algorithm starts by creating a random population of GP individuals and a random liquid. The evolutionary process is run for a fixed number of generations. The underlying algorithm for GP has been deeply described in \cite{koza1}. 

The modifications in liquid are also generation-based and usually they have a different rate compared to modifications performed for the GP individuals. From the numerical experiments we have deduced that the modifications in the liquid should not occur as often as the modifications within GP individuals. Thus an update of the liquid's items will be performed only after 5 generations of the GP algorithm. 

The updates for the liquid are as follows: at each generation of the liquid the following steps are repeated until the new \textit{LiquidSize} items are obtained: with a probability $p_{\mathrm{insert}}$ generate an offspring made up of a single terminal (see the Insertion operator, Sect. \ref{insertion}). With a probability $1-p_{\mathrm{insert}}$ randomly select two parents. The parents are recombined in order to obtain an offspring (see Sect. \ref{recombination}). 
The offspring enters the liquid of the next generation.

A basic form of elitism is employed by the GP algorithm: the best so far GP individual along with the current state of the liquid is saved at each generation during the search process. The best individual will provide solution of the problem.

\subsection{Complexity} \label{complexity}

A very important aspect of the GP techniques is the time complexity of the procedure used for computing the fitness of the newly created individuals.

The complexity of that procedure for the standard GP is $O(m*g)$, where $m$ is the number of fitness cases and $g$ is average number of nodes in the GP tree.

By contrast, the complexity of generating (by insertion or recombination) an individual in the liquid is only $O(m)$,  because the liquid's item is generated by traversing an array of size $m$ only once. The length of an item in the liquid is $m$.

Clearly, the use of the liquid could generate a small overhead of the LSGP when compared to the standard GP. Numerical experiments show (running times not presented due to the space limitation) that LSGP is faster than the standard GP, because the liquid part is considerably faster than the standard GP and many combinations performed in the liquid could lead to perfect solutions. However, it is very difficult to estimate how many generations we can run an LSGP program in order to achieve the same complexity as GP. This is why we restricted GP and LSGP to run the same number of generations.

\section{Numerical Experiments}
\label{exp}

Several numerical experiments using LSGP are performed in this section by using the even-parity problem. The Boolean even-parity function of $k$ Boolean arguments returns \textbf{T} 
(\textbf{True}) if an even number of its arguments are \textbf{T}. Otherwise 
the even-parity function returns \textbf{NIL} (\textbf{False}) \cite{koza1}.

\subsection{Experimental Setup}

General parameter of the LSGP algorithm are given in Table \ref{tab1}.

\begin{table*}[htbp]
\caption{General parameters of the LSGP algorithm}
\label{tab1}
\begin{center}
\begin{tabular}
{p{220pt}p{110pt}}
\hline
\textbf{Parameter}& 
\textbf{Value} \\
\hline
Liquid's insertion probability& 
0.05 \\
GP Selection& 
Binary Tournament \\
Terminal set for the liquid& 
Problem inputs \\
Liquid size& 
2 * Number of inputs \\
Terminal set for the GP individuals& 
Liquid's items \\
Number of GP generations before updating the liquid& 
5 \\
Maximum GP tree height & 
12 \\
Number of runs& 
100 (excepting for the even-8-parity)\\
\hline
\end{tabular}
\end{center}
\end{table*}

For the even-parity problems we use the set of functions (for both liquid and GP trees) $F$ = \{AND, OR, NAND, NOR\} as indicated in \cite{koza1}.

\subsection{Summarized Results}
\label{results}

Summarized results of applying Liquid State Genetic Programming for solving the considered problem are given in Table \ref{tab2}. 

For assessing the performance of the LSGP algorithm in the case of even-parity problems we use the success rate metric(the number of successful runs over the total number of runs).

\begin{table*}[htbp]
\caption{Summarized results for solving the even-parity problem using LSGP and GP. Second column indicates the population size used for solving the problem. Third column indicates the number of generations required by the algorithm. The success rate for the GP algorithms is given in the fourth column. The success rate for the LSGP algorithms is given in the fifth column}
\label{tab2}
\begin{center}
\begin{tabular}
{p{40pt}p{40pt}p{55pt}p{60pt}p{70pt}}
\hline
Problem& 
Pop size& 
Number of generations& 
GP success rate ($\%$)& 
LSGP success rate ($\%$) \\
\hline
even-3& 
100& 
50&
42& 
93 \\
even-4& 
1000& 
50&
9& 
82 \\
even-5& 
5000& 
50&
7& 
66 \\
even-6& 
5000& 
500&
4& 
54 \\
even-7& 
5000& 
1000&
-& 
14 \\
even-8& 
10000& 
2000&
-& 
1 successful out of 8 runs\\
\hline
\end{tabular}
\end{center}
\end{table*}

Table \ref{tab2} shows that LSGP is able to solve the even-parity problems very well. Genetic Programming without Automatically Defined Functions was able to solve instances up to even-5 parity problem within a reasonable time frame and using a reasonable population. Only 8 runs have been performed for the even-8-parity due to the excessive running time. Note again that a perfect comparison between GP and LSGP cannot be made due to their different individual representation.

Table \ref{tab2} also shows that the effort required for solving the problem increases with one order of magnitude for each instance.

\section{Limitations of the Proposed Approach}

The main disadvantage of the proposed approach is related to the history of the liquid which is not maintained. We cannot know the origin of an item from the liquid unless we store it. We can also store the tree for each item in the liquid but this will lead to a considerable more memory usage and to a lower speed of the algorithm. Another possible way to have access to the liquid's history is to store the entire items ever generated within the liquid and, for each item, to store pointers to its parents. This is still not very efficient; however, it is better than storing the entire tree, because, in the second case, the subtree repetition is avoided.

\section{Conclusions and Further Work}
\label{conclusions}

A new evolutionary technique called Liquid State Genetic Programming has been proposed in this paper. LSGP uses a special, dynamic memory for storing and manipulating the inputs.

LSGP has been used for solving several difficult problems. In the case of even-parity, the numerical experiments have shown that LSGP was able to evolve very fast a solution for up to even-8 parity problem. Note that the standard GP evolved (within a reasonable time frame) a solution for up to even-5 parity problem \cite{koza1}.

Further effort will be spent for improving the proposed Liquid State Genetic Programming technique. For instance, the speed of the liquid can be greatly improved by using Sub-machine Code GP \cite{poli1}, \cite{poli2}. Many aspects of the proposed LSGP technique require further investigation: the size of the liquid, the frequency for updating the liquid, the operators used in conjunction with the liquid etc.

The proposed LSGP technique will be also used for solving other symbolic regression, classification \cite{koza1} and dynamic (with inputs being variable in time) problems.

\end{document}